\documentclass[letterpaper]{article} 
\usepackage{aaai24}  
\usepackage{times}  
\usepackage{helvet}  
\usepackage{courier}  
\usepackage[hyphens]{url}  
\usepackage{graphicx} 
\usepackage{natbib}  
\usepackage{caption} 
\usepackage{algorithm}
\usepackage{listings}

\usepackage{algpseudocode}
\usepackage{tabularx, booktabs}
\usepackage{amsmath}
\usepackage{amssymb}
\usepackage{multirow}

\usepackage{newfloat}
\usepackage{listings}
\DeclareCaptionStyle{ruled}{labelfont=normalfont,labelsep=colon,strut=off} 
\floatstyle{ruled}
\newfloat{listing}{tb}{lst}{}
\floatname{listing}{Listing}
\title{Enhancing Sentiment Analysis Results through Outlier Detection Optimization }
\author{
    Yuetian Chen\textsuperscript{\rm 1}, Mei Si\textsuperscript{\rm 2}
    }
\affiliations{
    \textsuperscript{\rm 1}\textsuperscript{\rm 2}Rensselaer Polytechnic Institute\\

    110 8th street, \\
    Troy, New York 12180\\
    cheny63@rpi.edu, sim@rpi.edu
}

\usepackage{bibentry}

\begin{document}

\maketitle
\begin{abstract}
When dealing with text data containing subjective labels like speaker emotions, inaccuracies or discrepancies among labelers are not uncommon. Such discrepancies can significantly affect the performance of machine learning algorithms. This study investigates the potential of identifying and addressing outliers in text data with subjective labels, aiming to enhance classification outcomes. We utilized the Deep SVDD algorithm, a one-class classification method, to detect outliers in nine text-based emotion and sentiment analysis datasets. By employing both a small-sized language model (DistilBERT base model with 66 million parameters) and non-deep learning machine learning algorithms (decision tree, KNN, Logistic Regression, and LDA) as the classifier, our findings suggest that the removal of outliers can lead to enhanced results in most cases. Additionally, as outliers in such datasets are not necessarily unlearnable, we experienced utilizing a large language model -- DeBERTa v3 large with 131 million parameters, which can capture very complex patterns in data. We continued to observe performance enhancements across multiple datasets.

\end{abstract}

\section{Introduction}

When working on tasks that involve analyzing emotions and sentiments in text, it is common for different labelers to disagree on the labels. This is because these tasks are more subjective and context-dependent than other tasks, such as question answering or reference resolution. For example, consider the task of emotion classification in Twitter data. Tweets can be about a variety of topics, including personal experiences, daily life, reactions to global events, or just random thoughts. Because of this, tweets often combine facts with emotions, which makes it difficult to determine the primary emotion being expressed. As a result, different labelers may disagree on the label for a given tweet. This ambiguity and inconsistency can make it difficult to build accurate emotion analysis models.



\begin{figure*}[h]
    \centering
    \includegraphics[width=\linewidth]{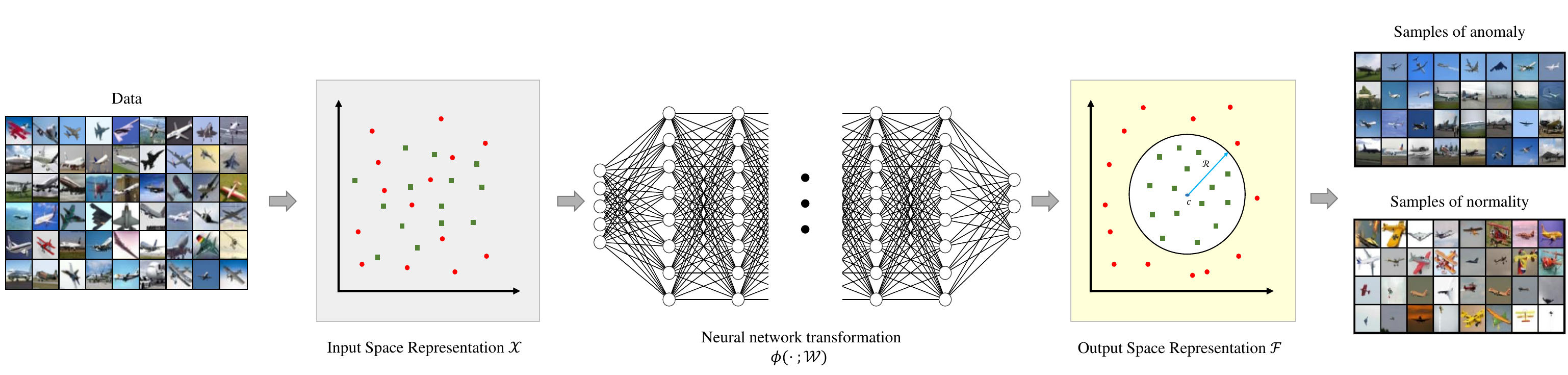}
    \caption{The Overview of the Application of Deep SVDD~\cite{ruff2018deep} in CIFAR 10}
    \label{fig:deepsvdd}
\end{figure*}

The examples in Table~\ref{tab:anomalie-example} are taken from the \texttt{mteb/emotion} dataset~\cite{saravia-etal-2018-carer}, which involves tweets labeled with emotions using hashtags like \texttt{\#depressed} and \texttt{\#joy}. In the possible outlier samples section, we notice that line 3144, despite discussing facts, is labeled as ``love". Similarly, line 7610, ``in a dam lake", could convey various emotions but is labeled as ``fear".

\begin{table}[htbp]
    \centering
    \begin{tabularx}{\columnwidth}{p{0.7cm}Xl}
    \toprule
    \multicolumn{3}{c}{\textbf{Normal Sample}} \\
    \midrule
    \textbf{Index} & \textbf{Text} & \textbf{Label} \\
    \midrule
    14921 & i start to feel unsure & 4 (``fear") \\
    8104 & i do really feel treasured by you too & 2 (``love") \\
    \midrule
    \multicolumn{3}{c}{\textbf{Possible Outlier Samples}} \\
    \midrule
    \textbf{Index} & \textbf{Text} & \textbf{Label} \\
    \midrule
    3144 & i have been told that these same vendors feel like they might end up supporting much more than just one more platform as Linux has many popular distribution releases these days & 2 (``love") \\
    7610 & in a dam lake & 4 (``fear") \\
    \bottomrule
    \end{tabularx}
    \caption{Examples of Possible Outliers in Emotion Labeling.}
    \label{tab:anomalie-example}
\end{table}

Researchers have taken note of the uncertainty that surrounds labeling emotion-related data. For instance, the EmotionX dataset, which involves emotion annotations for lines in the TV show ``Friends," gathered input from multiple annotators. Within this dataset, inconsistencies in agreement among annotators were evident, particularly regarding emotions like disgust and fear. These observations highlight the inherent ambiguity of assigning discrete emotion labels~\cite{shmueli2019socialnlp}. Similarly, another study~\cite{rashkin2018modeling} provided annotations for characters' needs, motives, and emotions within the ROCStories dataset~\cite{chen2019incorporating}. The agreement levels among labelers were found to be lowest for emotion labels compared to the other two types of labels. 

Inaccuracies and poor data quality can undermine the effectiveness of machine learning systems, leading to reduced accuracy and reliability. This study aims to enhance classification outcomes by excluding potentially incorrect or uncertain training data using a Deep SVDD~\cite{ruff2018deep} based approach. We hypothesize that using filtered datasets could expedite the learning process and yield improved results because the filtered datasets enable the machine learning algorithm to grasp the fundamental data features for the classification task. Outliers typically constitute only a minor fraction of the training data. Thus, by exclusively training on the inliers, we also anticipate the potential to boost classification results on the unrefined validation sets. 

Outliers within emotion labels don't necessarily equate to being incorrect or impossible to learn from. The presence of inconsistency or seemingly unconventional labels could merely highlight the intricate nature of language. Given that simpler machine learning models are more susceptible to outlier influence, we explored to assess the effectiveness of outlier filtering across three model types. The first category encompasses non-deep learning machine learning algorithms like Decision Trees~\cite{loh2011classification}, K-Nearest Neighbors (KNN)~\cite{altman1992introduction}, Logistic Regression~\cite{cramer2002origins}, and Linear Discriminant Analysis (LDA)~\cite{mika1999fisher}. The second category comprises a small language model, namely the DistilBERT base with 66 million parameters. The third category involves a substantial language model, DeBERTa v3 large with 131 million parameters, capable of comprehending highly intricate patterns in data.

Our study focuses on textual classification tasks, such as toxic comment identification, counterfactual detection, IMDb reviews, and sentiment analyses. We picked nine commonly used datasets. The details about these datasets are presented in Section~\ref{sec:data}. 

Our experiments indicate that eliminating outliers generally improves outcomes for non-deep learning machine learning algorithms and the small language model. However, with the large language model, performance improvements were evident in some datasets, while in others, it fared better using the unfiltered dataset. This outcome is in line with expectations. Our findings affirm that classifiers trained on the refined data (inliers) exhibit strong performance even on outlier data. This aligns with our original hypothesis that the refined datasets encapsulate more characteristic features for effective data classification, although certain valuable features relevant to specific data points might be overlooked.

On the other hand, considering computational resource usage, non-deep learning machine learning algorithms, and the small language model are significantly more economical to train and deploy. The DistilBERT base model, for instance, only needs 2G of memory. In contrast, the DeBERTa V3 model we employed demands 12G of memory. 


In Summary, the contributions of this work include:
\begin{itemize}
    \item We pioneered the application of Deep SVDD for data cleaning in textual domains and modified the Deep SVDD algorithm for this type of task. 
    \item Our experiments demonstrate that Deep SVDD is a versatile data-cleaning approach, capable of enhancing the performance of most classifiers across a wide range of scenarios.
\end{itemize}

\section{Related Work}

\subsection{Data Cleaning}
Cleaning data and spotting outliers are often performed in machine learning to boost model performance. Numerous methods have been proposed to tackle dirty data, including SampleClean~\cite{krishnan2015sampleclean}, ActiveClean~\cite{krishnan2016activeclean}, HoloClean~\cite{rekatsinas2017holoclean}, AlphaClean~\cite{krishnan2019alphaclean}, and CPClean~\cite{karlaš2020nearest}.

\textbf{SampleClean}~\cite{krishnan2015sampleclean} primarily focuses on aggregate query estimations on dirty data. For cleaning selected samples, it uses statistical techniques to yield bounded approximations of query results, particularly for SQL aggregates such as SUM and AVG. 
\textbf{ActiveClean}~\cite{krishnan2016activeclean} closely aligns with our work through its use of iterative data cleaning during model training. It mainly addresses convex loss models like linear regression and SVMs, offering convergence and error-bound assurances. 
\textbf{HoloClean}~\cite{rekatsinas2017holoclean} offers a comprehensive data cleaning approach, merging signals like integrity constraints, external dictionaries, and data statistics into a unified probabilistic model. 
\textbf{AlphaClean}~\cite{krishnan2019alphaclean} automates and optimizes data cleaning pipelines, breaking them down into granular repairs. 
Lastly, \textbf{CPClean}~\cite{karlaš2020nearest} delves into the ramifications of dirty data on machine learning models, introducing ``Certain Predictions" specifically for nearest neighbor classifiers. 

While each method discussed contributes valuable insights, our approach distinguishes itself by focusing on sentiment analysis and highlighting the effectiveness of targeted outlier detection and removal. Moreover, this step can operate independently of the classification model, making it compatible with any data classification pipeline.

\begin{figure*}[h]
    \centering
    \includegraphics[width=\linewidth]{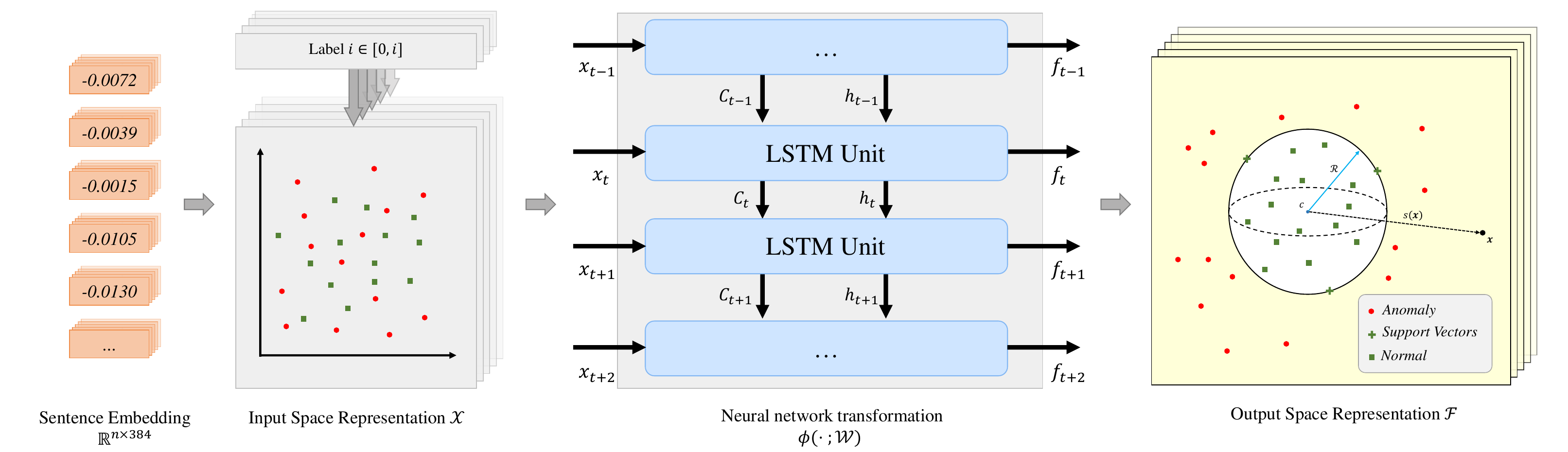}
    \caption{Pipeline for Processing Text Input}
    \label{fig:pipe}
\end{figure*}

\subsection{Deep Support Vector Data Description (Deep SVDD)}
Support Vector Data Description (SVDD) is mainly used for outlier detection~\cite{tax2004support}. It works by finding a sphere (in the feature space) that contains most of the data, aiming to encapsulate the majority of the target (normal) data, while outliers lie outside of this sphere. The primary objective of SVDD is to identify the smallest hypersphere, characterized by its center $\mathbf{c} \in \mathcal{F}_k$ and radius $R > 0$, which encompasses the majority of the data within the feature space $\mathcal{F}_k$. This is mathematically formulated as follows:

\begin{equation}
    \begin{aligned}
    \min_{R, \mathbf{c}, \xi} & \quad R^2 + \frac{1}{vn} \sum_i  \xi_i \\
    \text{s.t.} & \quad ||\phi_k(\mathbf{x}_i) - \mathbf{c}||^2_{\mathcal{F}_k} \leq R^2 + \xi_i, \\
    & \quad \xi_i \geq 0, \forall i
    \end{aligned}
    \label{eq:SVDD_eq}
\end{equation}

In this formulation, the slack variables $\xi_i \geq 0$ introduce flexibility in defining the boundary of the hypersphere. The hyperparameter $v \in (0, 1]$ modulates the balance between penalties $\xi_i$ and the hypersphere's volume. Data points situated outside this hypersphere, specifically where $||\phi_k(\mathbf{x}_i) - \mathbf{c}||^2 > R^2$, are classified as outliers.

Deep SVDD augments the SVDD approach by jointly training a neural network to project the input data into a minimum-volume hypersphere in the output space~\cite{ruff2018deep}. Figure~\ref{fig:deepsvdd} shows an example of using deep SVDD to identify outliers in CIFAR10~\cite{krizhevsky2009learning}. 

The core proposition of Deep SVDD is the synergy of data representation learning with the one-class classification goal, achieved through a neural network trained to project data into a compact hypersphere. Given the premise that a significant portion of training data $D_n$ is not an outlier, Deep SVDD aims to optimize the \textit{One-Class Deep SVDD} objective:

\begin{equation}
    \begin{aligned}
    \min_{\mathcal{W}} & \quad \frac{1}{n} \sum^{n}_{i=1} ||\phi(\mathbf{x}_i; \mathcal{W}) - \mathbf{c}||^2 + \frac{\lambda}{2} \sum^{L}_{\ell=1}||\mathbf{W}^\ell||^2_F
    \end{aligned}
    \label{eq:DeepSVDD_eq}
\end{equation}


To formalize the methodology, consider an input space $\mathcal{X} \subseteq \mathbb{R}^d$ and an associated output space $\mathcal{F} \subseteq \mathbb{R}^p$. Let $\phi(\cdot ; \mathcal{W}) : \mathcal{X} \xrightarrow{} \mathcal{F}$ represent a neural network, encapsulating $L \in \mathbb{N}$ hidden layers. The weight set $\mathcal{W} = \{\mathbf{W}^1, \dots, \mathbf{W}^L\}$ characterizes the network, with $\mathbf{W}^L$ being the weights for layer $\ell \in \{1, \dots, L\}$. Consequently, $\phi(\mathbf{x}; \mathcal{W}) \in \mathcal{F}$ describes the feature representation of $\mathbf{x} \in \mathcal{X}$, derived from network $\phi$ with parameter $W$, and therefore can map normal examples fall within, whereas mappings of outliers fall outside the hypersphere.

The One-Class Deep SVDD adopts a quadratic loss to penalize deviations in network representations, specifically, the distance between each representation $\phi(\mathbf{x}; \mathcal{W})$ and the center $c$ in the feature space $\mathcal{F}$. An additional network weight decay regularizer is incorporated, parameterized by $\lambda > 0$. Conceptually, the objective of One-Class Deep SVDD can be visualized as determining a hypersphere in the feature space with a minimal volume, centered at $c$. This optimization aims to contract the hypersphere by minimizing the average distance of all data representations from its center. The underlying rationale is to guide the neural network to discern and capture the predominant patterns or factors in the data. By penalizing deviations in the average distance of all data points and not merely outliers, the methodology aligns with the premise that the preponderance of training data pertains to a singular class.

\section{Adapting Deep SVDD for Identifying Outliers in Textual Data}

\subsection{Modifications}
In our proposed pipeline, illustrated in Figure~\ref{fig:pipe}, we introduced several modifications to adapt the original Deep SVDD for multi-class textual datasets. First, since Deep SVDD was initially designed for single-class data, we partitioned the multi-class data into distinct groups based on their labels. Each group, representing a specific class, underwent the Deep SVDD process independently to identify outliers. This approach preserves the unique characteristics of each class in subsequent processes.

Additionally, Deep SVDD was initially optimized for image-related tasks, primarily operating with image inputs and features extracted from CNN-based networks like LeNet~\cite{lecun1998gradient} and ResNet~\cite{he2015deep}. When adapting this algorithm for text inputs, we transformed the textual data into embeddings. For this purpose, we utilized the sentence transformers model~\cite{reimers2019sentencebert} `all-MiniLM-L6-v2' from the HuggingFace Transformer library. To further enhance feature extraction from the text, we incorporated a Long Short-Term Memory (LSTM) layer~\cite{graves2012long} after the embedding layer.

Finally, we constructed an outlier score. The main goal of Deep SVDD is two-fold: refining the network parameters $W$ iteratively and simultaneously minimizing the volume of a hypersphere in the output space $\mathcal{F}$. This hypersphere, defined by radius $R > 0$ and center $\mathbf{c} \in \mathcal{F}$, encompasses the data and serves as an anchor for detecting outliers in subsequent applications."

The outlier detection capability is quantified using an outlierness score, $s(x)$, for each data point $x \in \mathcal{X}$. Defined by the distance of the point to the hypersphere's center, the score is given as:

\begin{equation}
    \begin{aligned}
        s_{norm}(x) &= \frac{s(x) - min(S)}{max(S) - min(S)} \\
        s.t. \quad s(x) &= ||\phi(x; W^*) - c||^2 \\
        S &= \{s(x_i) | x_i \in \mathcal{X}\}
    \end{aligned}
\end{equation}

The value of the outlierness score is in the range of 0 to 1. A value being 1 means the data point is likely to be an outlier, while a value closer to 0 indicates that the data point aligns well with the main data distribution and is less likely to be an outlier. This score, generated by DeepSVDD, serves as a metric for determining the ``anomaliness" of each data point within the dataset.




\subsection{Steps for Training Deep SVDD}

\begin{algorithm}
\caption{Training and Inferencing with DeepSVDD}\label{SVDD_train}
\begin{algorithmic}[1]
\Require Datasets $\mathcal{X}$ with $n$ entries and $i$ classes
\Require normal\_class $\in \{0, \dots, i\}$


\State \# Step 1. Transform textual data to sentence embedding
\For{$ (x_j, y_j) \in X $}
    \State $ x_j $ = SentenceEmbedding($ x_j $) \Comment{s.t. $ x_j \in \mathbb{R}^{384} $}
    \State $ y_j $ = [$ y_j = k $]
\EndFor
\State 

\State \# Step 3. Initialization \& Pretraining
\State $\mathcal{R} = 0$
\State $\mathbf{c} = null$
\State $net = \phi(\cdot ; \mathcal{W})$
\State DeepSVDD = DeepSVDD($\mathcal{R}, \mathbf{c}, net$)
\State $\phi$ = BuildDecoder($ \phi $)
\State $\phi$.pretrain($\mathcal{X}$)  \Comment{$\ell = mean(  \phi(x_{batch}; \mathcal{W}) $ - $ x_{batch} )^2$}
\State
\State \# Step 4. Train SVDD
\State DeepSVDD.$\phi$ = LoadEncoder($\phi$)
\State DeepSVDD.train($\mathcal{X}$) \Comment{update $\mathcal{R}, \mathbf{c}, net$}
\State 

\State \# Step 5. Inference: calculate the outlierness score
\For{each entry $ x_j $ in $ X $}
    \State $ S[j] = s(x_j) = ||\phi(x; W^*) - c||^2 $
\EndFor

\For{each entry $ x_j $ in $ X $}
    \State $ s_{norm}(x_j) $ = $\frac{s(x_j) - \min(S)}{\max(S) - \min(S)}$
\EndFor

\end{algorithmic}
\end{algorithm}


The steps for identifying outliers are outlined in Algorithm~\ref{SVDD_train}. To facilitate the training process and ensure that the network converges efficiently, following~\cite{ruff2018deep}, we introduce a pretraining phase using an autoencoder. This autoencoder consists of two main parts: an encoder and a decoder. The encoder's role is to convert input text into a compact, fixed-size representation. In contrast, the decoder aims to reconstruct the initial input using this condensed representation.


Once the autoencoder completes its training phase, we transfer the weights from the encoder to initialize the Deep SVDD network. This step ensures that the Deep SVDD model inherits the feature-detection capabilities and representations learned during the autoencoder phase, providing a robust foundation. Then in step 4, the Deep SVDD model is trained as in~\cite{ruff2018deep}. Finally, in step 5, we can use the trained model to infer an outlierness score for each data point.



\section{Experimental Setup}

\subsection{Model and Training Configuration}

We conducted an exploration to evaluate the effectiveness of outlier filtering across three different model types. The first category encompasses non-deep learning machine learning algorithms, such as Decision Tree~\cite{loh2011classification}, K-Nearest Neighbors (KNN)~\cite{altman1992introduction}, Logistic Regression~\cite{cramer2002origins}, and Linear Discriminant Analysis (LDA)~\cite{mika1999fisher}. The second category involves a smaller language model, specifically the DistilBERT base with 66 million parameters. The third category includes a more substantial language model, DeBERTa v3 large with 131 million parameters. For the small and large language models, we utilized two transformer models from the HuggingFace library: `distilbert-base-uncased'~\cite{sanh2020distilbert} and `microsoft/deberta-v3-large'~\cite{he2021deberta}, respectively.

In terms of dataset splits, all datasets adhere to their default split configurations. In cases where datasets lack predefined train/validation/test splits, a consistent strategy is employed: 10\% of the data is set aside for testing, while the remaining 90\% is allocated for training. 


\begin{figure*}[h]
    \centering
    \includegraphics[width=\linewidth]{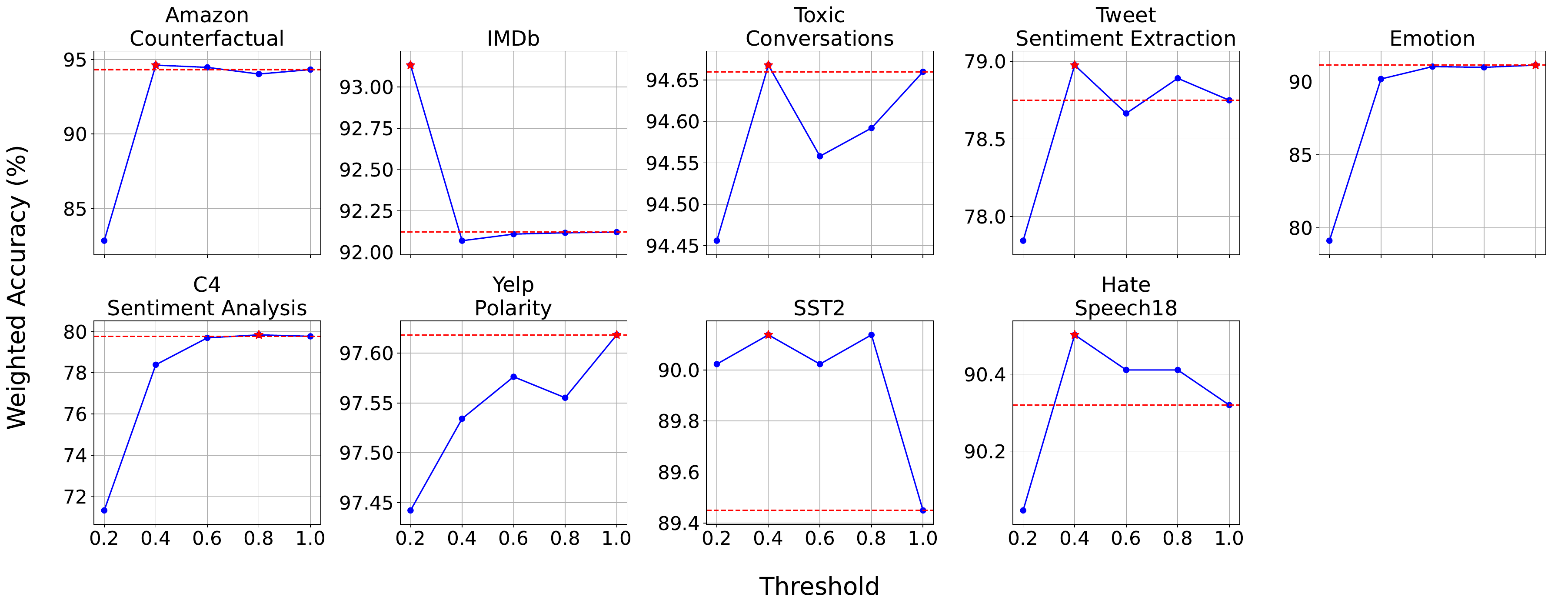}
    \caption{Evaluation Result on \texttt{distilbert-base-uncased}}
    \label{fig:distilbert-eval}
\end{figure*}

\begin{figure*}[h]
    \centering
    \includegraphics[width=\linewidth]{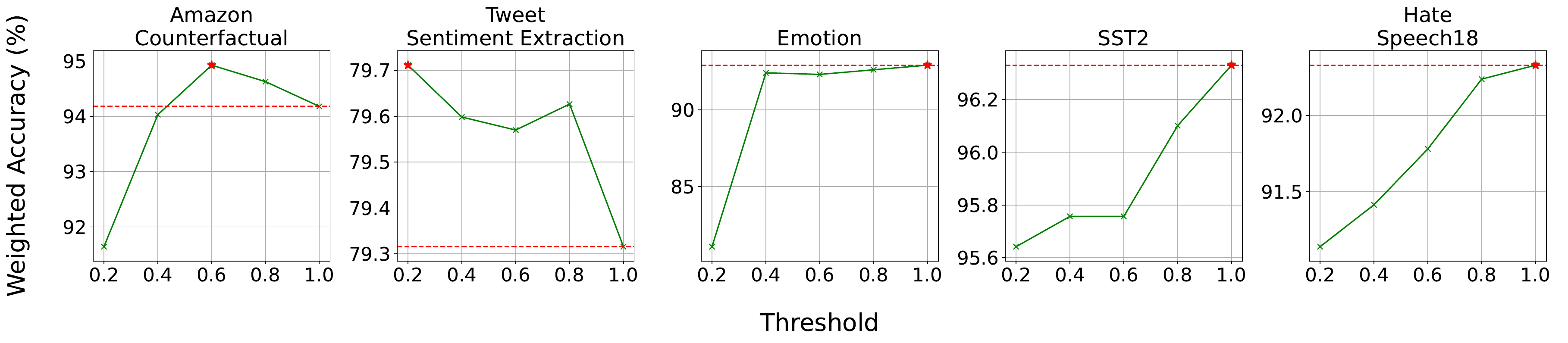}
    \caption{Evaluation results on \texttt{deberta-v3-large}}
    \label{fig:deberta-eval}
\end{figure*}

\subsection{Data Description}
\label{sec:data}
To empirically evaluate our approach, we utilized a diverse set of sentiment analysis datasets, each with distinct emphases on various domains and levels of granularity. These datasets encompass sentiments across a myriad of contexts, spanning from product reviews on renowned platforms such as Amazon and Yelp to nuanced emotions embedded within textual content.



\textbf{Amazon Counterfactual Classification:}
The dataset~\cite{oneill2021i} aims to classify statements as counterfactual or not, which are hypothetical scenarios contrary to reality. It contains 3,253 instances in the \emph{not-counterfactual} class and 765 instances in the \emph{counterfactual} class. The dataset's imbalance, with fewer counterfactual instances, can pose challenges for building robust models.

\textbf{IMDb Classification:}
The \texttt{imdb} dataset~\cite{maas-etal-2011-learning} is a balanced resource for sentiment analysis, including 12,500 \emph{negative} and 12,500 \emph{positive} movie reviews. Originating from IMDb, this dataset covers diverse opinions on various films, genres, and cinematic aspects, making it a benchmark for sentiment analysis models.

\textbf{Toxic Conversations Classification:}
The dataset~\cite{jigsaw-toxic-comment-classification-challenge} targets the identification of toxic comments, crucial for maintaining healthy online interactions. It consists of 46,035 \emph{non-toxic} entries and 3,965 \emph{toxic} entries. The dataset's imbalanced nature mirrors real-world scenarios, presenting challenges in detecting harmful content while minimizing false alarms.

\begin{table*}[htbp]
    \centering
    \begin{tabularx}{\linewidth}{Xccccc}
        \toprule
        \textbf{Dataset} & \textbf{Threshold} & \textbf{Decision Tree} & \textbf{KNN} & \textbf{Logistic Regression} & \textbf{LDA} \\
        \midrule
        \multirow{5}{*}{Amazon Counterfactual Classification} 
        & 0.2 & 84.925\% & 86.418\% & 82.537\% & \textbf{77.015\%} \\
        & 0.4 & 89.104\% & 86.866\% & 87.612\% & 68.955\% \\
        & 0.6 & \textbf{90.597\%} & 85.821\% & 89.552\% & 64.776\% \\
        & 0.8 & 90.000\% & \textbf{85.970\%} & \textbf{90.299\%} & 63.731\% \\
        & 1.0 & 89.552\% & 84.925\% & \textbf{90.299\%} & 63.284\% \\
        \midrule
        \multirow{5}{*}{IMDb Classification} 
        & 0.2 & \textbf{70.648\%} & 66.096\% & 88.288\% & \textbf{51.752\%} \\
        & 0.4 & 70.632\% & \textbf{66.184\%} & 88.284\% & 51.708\% \\
        & 0.6 & 70.280\% & 66.148\% & 88.304\% & 51.700\% \\
        & 0.8 & 70.464\% & 66.140\% & 88.312\% & 51.692\% \\
        & 1.0 & 70.292\% & 66.152\% & \textbf{88.316\%} & 51.687\% \\
        \midrule  
        \multirow{5}{*}{Toxic Conversation Classification} 
        & 0.2 & \textbf{92.458\%} & 92.108\% & \textbf{92.574\%} & \textbf{54.516\%} \\
        & 0.4 & 91.996\% & 92.142\% & 93.160\% & 54.502\% \\
        & 0.6 & 91.964\% & \textbf{92.148\%} & 93.268\% & 54.485\% \\
        & 0.8 & 91.896\% & \textbf{92.148\%} & 93.284\% & 54.493\% \\
        & 1.0 & 91.924\% & \textbf{92.148\%} & 93.274\% & 54.478\% \\
        \midrule
        \multirow{5}{*}{Emotion Classification} 
        & 0.2 & 60.700\% & 51.550\% & 57.350\% & 46.050\% \\
        & 0.4 & \textbf{60.800\%} & 54.050\% & 60.600\% & 50.000\% \\
        & 0.6 & 59.450\% & \textbf{54.300\%} & \textbf{61.050\%} & \textbf{51.350\%} \\
        & 0.8 & 60.000\% & 54.050\% & \textbf{61.050\%} & 50.950\% \\
        & 1.0 & 59.050\% & 54.200\% & \textbf{61.050\%} & 51.000\% \\
        \midrule
        \multirow{5}{*}{Hate Speech18} 
        & 0.2 & 85.479\% & 87.763\% & 88.584\% & 73.973\% \\
        & 0.4 & 85.479\% & 87.763\% & \textbf{88.678\%} & 75.251\% \\
        & 0.6 & 85.297\% & 87.763\% & 88.584\% & \textbf{75.799\%} \\
        & 0.8 & \textbf{85.571\%} & 87.763\% & 88.584\% & 74.886\% \\
        & 1.0 & 86.027\% & 87.763\% & 88.584\% & 75.068\% \\
        \bottomrule
    \end{tabularx}
    \caption{Weighted Accuracy Evaluation using Non-Deep Learning Models}
    \label{tab:weighted-accuracy-evaluation}
\end{table*}

\begin{table*}[htbp]
    \centering
    \begin{tabularx}{\linewidth}{Xccccc}
        \toprule
        \textbf{Dataset} & \textbf{Threshold} & \textbf{Data Coverage (\%)} & \multicolumn{3}{c}{\texttt{distilbert-base-uncased}} \\
        & & & \textbf{Inlier} & \textbf{Outlier} & \textbf{Weighted} \\
        \midrule
        \multirow{2}{*}{Amazon Counterfactual Classification} & 0.6 & 65.620\% & \textbf{94.805\%} & 89.313\% & \textbf{94.627\%} \\
         & 1.0 & 100\% & 94.434\% & 92.366\% & 94.328\% \\
        \midrule
        \multirow{2}{*}{Hate Speech18} & 0.4 & 95.235\% & \textbf{92.680\%} & 73.600\% & \textbf{90.502\%} \\
         & 1.0 & 100\% & 92.371\% & 74.400\% & 90.319\% \\
        \bottomrule
    \end{tabularx}
    \begin{tabularx}{\linewidth}{Xccccc}
        \toprule
        \textbf{Dataset} & \textbf{Threshold} & \textbf{Data Coverage (\%)} & \multicolumn{3}{c}{\texttt{deberta-v3-large}} \\
        & & & \textbf{Inlier} & \textbf{Outlier} & \textbf{Weighted} \\
        \midrule
        \multirow{2}{*}{Amazon Counterfactual} & 0.6 & 94.027\% & \textbf{96.805\%} & 89.312\% & \textbf{94.925\%} \\
         & 1.0 & 100\% & 94.804\% & 90.840\% & 94.179\%\\
        \midrule
        \multirow{2}{*}{Hate Speech18} & 0.4 & 95.235\% & 91.649\% & 89.600\% & 91.416\% \\
         & 1.0 & 100\% & \textbf{91.959} \% & 90.400\% & \textbf{91.780\%} \\
        \midrule
    \end{tabularx}
    
    \caption{Accuracy Evaluation using Deep Learning Models}
    \label{tab:anomalie-example-weight-accuracy}
\end{table*}

\textbf{Tweet Sentiment Extraction:}
The \texttt{tweet-sentiment-extraction} dataset~\cite{tweet-sentiment-extraction} showcases sentiment distribution in tweets with three classes: \emph{negative} (7,781 entries), \emph{neutral} (11,118 entries), and \emph{positive} (8,582 entries).

\textbf{Emotion Classification:}
The \texttt{emotion} dataset~\cite{saravia-etal-2018-carer} contains text labeled with six emotions: \emph{joy} (5,362), \emph{sadness} (4,666), \emph{anger} (2,159), \emph{fear} (1,937), \emph{love} (1,304), and \emph{surprise} (572), offering a nuanced sentiment classification task.

\textbf{C4 Sentiment Analysis:}
Derived from C4~\cite{raffel2020exploring}, this dataset includes 250k entries with sentiments: \emph{positive} (52,906), \emph{negative} (13,527), and \emph{neutral} (187,673). Annotated by GPT3.5, it challenges models with machine-generated sentiment annotations.

\textbf{Yelp Polarity:}
The \texttt{yelp\_polarity} dataset~\cite{zhangCharacterlevelConvolutionalNetworks2015} simplifies Yelp reviews into \emph{positive} and \emph{negative} sentiments (both 280,000), serving as a strong foundation for binary sentiment classification models across diverse domains.

\textbf{SST2:}
From Stanford Sentiment Treebank~\cite{socher-etal-2013-recursive}, the \texttt{sst2} dataset includes \emph{1} (37,569) and \emph{0} (29,780) sentiment classes, spanning movie reviews with varying levels of sentiment expression.

\textbf{Hate Speech18:}
The \texttt{hate\_speech18} dataset~\cite{gibert2018hate} categorizes content into \emph{noHate} (8,551), \emph{hate} (1,079), \emph{relation} (153), and \emph{idk/skip} (66) classes, facilitating analysis of diverse online behaviors.

\section{Results and Discussion}
\begin{figure}
    \centering
    \includegraphics[width=.9\columnwidth]{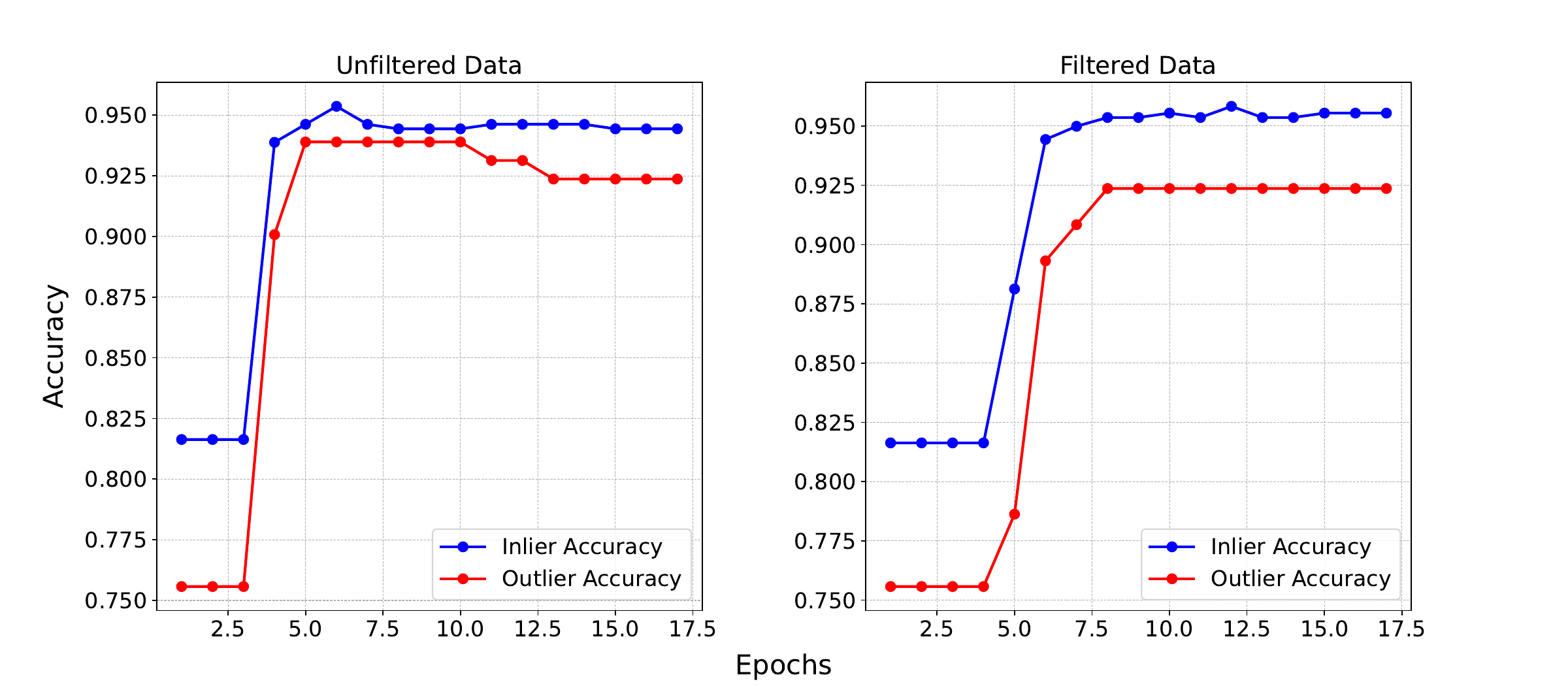}
    \caption{Comparison of inlier and outlier accuracies over epochs}
    \label{fig:inlier_outlier_comparision}
\end{figure}

Central to our experimental design is the concept of the outlier removal threshold. This threshold operates on the outlier scores produced by Deep SVDD, which range from 0 to 1. By setting a specific threshold, we filter out data entries with outlier scores exceeding that threshold. For instance, a strict threshold of 0.2 implies rigorous filtering, allowing only data with scores up to 0.2 to be retained. Conversely, a threshold of 1 is the most lenient, signifying no data removal since all scores fall within the 0 to 1 range. Through our experiments, we explored a variety of thresholds to discern the optimal balance between rigorous filtering and information retention.

\subsection{DistilBERT Base}

Figure \ref{fig:distilbert-eval} depicts the evaluation performance of the small language model, DistilBERT base, across different thresholds. For all these experiments, the model was trained using the filtered data, and tested at the unfiltered validation data. Each plot corresponds to results for a dataset, with the red dashed line denoting the baseline accuracy at threshold=1. The red star symbol on each plot signifies the best performance across thresholds. The consistent observation is that the 0.6-0.8 threshold range often yields balanced performance. Extremely low thresholds like 0.2 tend to remove crucial data, potentially leading to information loss. Conversely, thresholds close to 1 allow potential outliers to infiltrate the training data.

To further study how the outlier filtering process affects learning outcomes, we created Figure \ref{fig:inlier_outlier_comparision} which presents a comparative analysis of inliers and outliers within the ``Amazon Counterfactual Classification" dataset evaluated by the DistilBERT base model. The distinct trajectories of inliers and outliers shed light on the efficacy of our data-cleaning methodology. Outliers consistently underperformed, with longer convergence timelines, regardless of whether the training data was cleaned. Post data cleaning, the performance disparity between inliers and outliers widened, with inliers registering a significant uptick in accuracy.


\subsection{DeBERTa v3 Large}
Figure \ref{fig:deberta-eval} presents the evaluation results utilizing the DeBERTa v3 large model. Given resource and computational limitations, we focused on a subset of datasets for this heavyweight model. An evident trend is the subtle variation in performance improvements following data cleaning compared to the DistilBERT base model. This observation suggests that larger models like DeBERTa are capable of learning complex patterns even from ambiguous data. 

Upon closer examination of the results, several intriguing patterns come to light. In the Amazon Counterfactual dataset, there's a clear performance boost when transitioning from a threshold of 0.2 to 0.6. Similarly, the Tweet Sentiment Extraction and SST2 datasets exhibit analogous trends, where strict filtering isn't consistently linked to the worst performance, and the baseline doesn't always equate to the best results. Instances arise where the data cleaning process doesn't yield anticipated enhancements.

Contrary to our hypothesis, the Emotion dataset experiences a performance dip at a threshold of 0.4. This could stem from factors like the dataset's relative cleanliness or the stringent filtering inadvertently discarding semantically valuable data points.

Moreover, the Hate Speech18 dataset showcases a consistent performance upswing as the threshold escalates, indicating that larger models can gain from a precisely tuned data cleaning procedure. These observations underscore the intricate interplay between data characteristics, model complexity, and the efficacy of data cleaning thresholds.

Table \ref{tab:anomalie-example-weight-accuracy} furnishes a comprehensive analysis of our data-cleaning approach when applied to state-of-the-art deep learning models. The DeBERTa v3 large, emblematic of sophisticated deep learning architectures, exhibits remarkable resilience. Its performance, particularly in datasets like Amazon Counterfactual, remains steadfastly close to the baseline. However, a more granular observation reveals that even such advanced models stand to gain from judicious data cleaning. In the Hate Speech18 dataset, for instance, the inliers' performance under the DeBERTa v3 large model at a threshold of 0.4 (92.680\%) slightly outpaces that at threshold 1.0 (92.371\%), suggesting our method's potential to refine even already optimized datasets. 

However, there were instances where the data cleaning process didn't yield the expected improvements. The Emotion dataset, for instance, showed a drop in performance with the threshold set at 0.4, which is a counter-example to our hypothesis. This could be attributed to various factors: perhaps the dataset was already relatively clean, or the more rigorous filtering inadvertently removed some semantically-rich data points.

Another intriguing observation is seen in the Hate Speech18 dataset. Here, the data cleaning process demonstrated a consistent uptick in performance as the threshold increased, with the model achieving its peak performance at the highest threshold. This suggests that in this particular dataset, the larger models, with their inherent resilience to noise and outliers, are adept at handling the outliers present. The data filtering process, which reduces the sample size, appears to be a detriment rather than an advantage in this context, indicating that the model performs best when trained on the complete dataset without any exclusions.

\subsection{Traditional Machine Learning Models}
Table \ref{tab:weighted-accuracy-evaluation} provides a comprehensive analysis of our approach's effectiveness across traditional machine learning models. A consistent trend emerges: Logistic Regression consistently outperforms other models, either matching or exceeding the baseline across all datasets and thresholds. This suggests Logistic Regression's adaptability and resilience to the alterations in data caused by our cleaning process.

Conversely, Linear Discriminant Analysis (LDA) displays heightened sensitivity. For instance, in the ``Amazon Counterfactual Classification" dataset, LDA's performance significantly drops as the threshold increases, plunging from 77.015\% at a threshold of 0.2 to 63.284\% at a threshold of 1. This stark decline underscores LDA's dependency on the complete data distribution, which our cleaning procedure might have disrupted.

Interestingly, the ``Emotion Classification'' dataset demonstrates minimal fluctuation across models and thresholds, suggesting intrinsic cleanliness or a scarcity of pronounced outliers. This underscores the importance of adaptive threshold tuning, emphasizing the need to align with each dataset's unique characteristics and distribution.

Across all datasets, our data-cleaning approach consistently ushered in accuracy improvements. Such performance augmentation affirms the efficacy of our approach in eliminating noise and refining classifier training. However, there were instances, albeit rare, where our cleaning regimen either maintained the accuracy or led to marginal reductions. Such outcomes could be attributed to the inherent cleanliness of the dataset or the inadvertent elimination of rare but crucial data points during the cleaning process.



\section{Future Work}
While our work shows exciting potential in cleaning textual data using Deep SVDD, there are still ways to improve and explore. We are particularly interested in the following directions:

\paragraph{Dynamic Thresholding} Currently, our methodology leverages a static threshold to demarcate outliers based on the normalized outlierness score. While effective in many scenarios, a one-size-fits-all threshold might not always capture the nuances of diverse datasets. A more adaptive approach, such as dynamic thresholding, could calibrate based on the intrinsic characteristics of the data, considering factors like its distribution or inherent variability. By allowing the model to adjust its threshold based on the data it encounters, we can potentially achieve more precise outlier detection, reducing false positives and negatives.

\paragraph{Incorporation of Attention Mechanisms}
Attention mechanisms are extensively used in natural language processing tasks, allowing models to spotlight important segments of text. Introducing these mechanisms could be valuable in outlier detection. By assigning weights to various parts of the text based on their significance, attention mechanisms might enhance the precision of our detection. This could ensure that even subtle outliers are identified without being overshadowed by more prevalent patterns.

\paragraph{Multi-modal Outlier Detection} The confluence of textual data with other modalities poses unique challenges and opportunities. An intriguing research avenue would be to extend the Deep SVDD framework to handle multi-modal data, examining how outliers manifest in such composite datasets. This would require a thorough understanding of feature fusion techniques and the development of novel architectures that can seamlessly integrate information across modalities.

\section{Conclusion}
Our study was motivated by the pivotal role of data cleanliness in machine learning, especially within sentiment analysis. We adapted the deep SVDD model to systematically detect and eliminate outliers from text datasets. Our findings showcased a consistent accuracy boost for smaller models after outlier removal, with the optimal threshold falling between 0.6 and 0.8. While larger models are naturally more robust to outliers, they still glean some advantages from the cleaning process, albeit to a lesser degree. Essentially, our research highlights the significant influence of data cleanliness on model performance, introducing a scalable technique with tangible advantages.

\section*{Acknowledgement}

We would like to extend our heartfelt gratitude to Professor Yu Lei for his invaluable guidance throughout this research. His insights and expertise were instrumental in the shaping and success of this study. Additionally, we appreciate the generous computational resources provided by him, which were crucial during the evaluation phase of our work.

\section*{Appendix}
\subsection{Deployment Solution}
We have seamlessly integrated our outlier detection solution into our own server, offering it as a streamlined service. Given the model-independence of our framework, users can, with minimal effort, train and induct Deep SVDD models by providing dataset information, which encompasses data point content and optional labeling information. With a single line of code, outlierness scores for thorough analysis and filtering can be obtained. It's worth noting that the total amount of data or the amount of data for a single label should be at least greater than 500 samples. This is essential to ensure proper training of DeepSVDD, reliable inference, and to mitigate the risk of underfitting the model after filtering due to insufficient data. As for deployment considerations, our solution, while not being computation-intensive, mandates the use of a dedicated GPU with a minimum of 8 GB of memory when employing the default LSTM architecture for the neural network transformation. Batch processing scenarios would ideally require at least 16 GB of RAM. It's pivotal to ensure that the distribution of the training data is kept consistent with that of the test data. Any significant disparities could amplify the false positives in outlier detection. Further, the sensitivity threshold for detecting outliers, which is set at 0.60 for our showcase, may require fine-tuning based on specific use cases and the tolerable false positive rate. Post-deployment, the significance of engaging in routine retraining of both the sentiment model and the outlier detection mechanism cannot be overstated, especially to accommodate the shifting dynamics of customer sentiments and linguistic subtleties. Lastly, embedding a feedback mechanism for end-users to highlight any perceived inaccuracies in sentiment predictions not only aids in model refinement but also fosters trust and transparency between the service and its users.

\section{Usage Example}

To elucidate the practical application of our framework, consider a dataset represented by data $ x $ and labels $ y $. The \texttt{train} set is a collection of indices that delineate the training subset of this data. For context, imagine $ x $ as a series of text reviews, with $ y $ being their associated sentiment labels.

The primary operation involves the \texttt{autoFilter.filter\_data} method. This method computes the outlierness scores and filters the data based on a designated threshold, yielding a refined set of training indices. 

\begin{lstlisting}[language=Python, caption=usage examples]
# Pseudo-code
function filter_data(train_indices, test_indices, data, labels, threshold):
    compute outlierness scores for data
    filter data based on threshold
    return updated train_indices
end function

# Actual usage
train = autoFilter.filter_data(train, test, x, y, threshold=0.6)
\end{lstlisting}

In this setup, \texttt{train} is the refreshed set of indices for the training subset. The \texttt{threshold} parameter, set to 0.6 here, steers the sensitivity for outlier identification.

To offer additional perspective:

\begin{itemize}
    \item $ x $ could be sequenced like [``The product is good", ``A terrible experience", \dots].
    \item $ y $ might be sentiments such as [Positive, Negative, \dots].
\end{itemize}

Post-execution, \texttt{train} encompasses the indices of reviews that align closely with the dataset's overarching patterns.

It's imperative to note that while indices represent data points in this illustration, our framework accommodates more intricate representations. The \texttt{threshold} can also be adjusted to cater to application-specific needs. A loftier value might be lenient, whereas a lower one could be more rigorous in outlier filtration.

Our chief design ambition is to simplify the interplay between advanced outlier detection algorithms and raw data. This ensures users can harness DeepSVDD's prowess without immersing themselves in its complexities. The outcome is a meticulously curated training dataset, primed to heighten the precision and trustworthiness of ensuing sentiment analysis models.

Based on this framework, some potential application would be:

\textbf{1. Financial Transaction Review}: 
In the banking and finance sector, ensuring the accuracy of sentiment analysis on customer feedback is pivotal. By filtering out potential outliers from the training dataset, our framework can help in creating more robust models for analyzing customer reviews. For instance, a bank can accurately gauge customer sentiments regarding a new digital banking feature, ensuring that the sentiment analysis is not skewed by non-representative feedback. This refined understanding can then guide product improvements and enhance customer satisfaction.

\textbf{2. Healthcare Patient Feedback Analysis}: 
Hospitals and healthcare providers continually seek patient feedback to improve services. Given the sensitive nature of healthcare, it's crucial to have an accurate sentiment analysis model. By utilizing our framework, healthcare institutions can filter out anomalies from patient feedback datasets, ensuring that the sentiment models are trained on representative data. For example, a hospital can get a clear sentiment analysis on patient feedback about a new telehealth service, ensuring that the insights drawn are based on the majority sentiment, free from potential outliers that might not reflect the broader patient experience.

These applications underscore the versatility and significance of our outlier detection framework in refining sentiment analysis across diverse sectors.

\section{Real-world application}

The datasets we engaged for evaluation are not just theoretical constructs but have propelled advancements in multifarious arenas, especially in online content moderation, sentiment analysis, and counterfactual reasoning.

The \textit{Amazon Counterfactual Classification} dataset, scrutinized by~\cite{bottou2013counterfactual}, has been pivotal for discerning the ramifications of counterfactual reasoning on model resilience and adaptability. This dataset underpins how models, when exposed to hypothetical scenarios or alternate outcomes, can be honed to exhibit enhanced robustness. Such an approach becomes indispensable in platforms where users frequently broach or debate "what-if" situations.

On the other hand, the \textit{Toxic Conversations Classification} dataset has emerged as a linchpin for fostering healthier digital discourse. With the insights from ~\cite{saveski2021structure}, it's evident that this dataset holds the potential to reshape social media platforms into spaces of constructive dialogue. By tapping into the intrinsic structure of online conversations, imminent toxic interactions can be identified and possibly redirected. Its utility doesn't stop here. As illustrated by~\cite{kim2019topic}, platforms like Quora can harness this dataset to sieve out malignant content. Furthermore, when the BERT model is trained on this treasure trove of data, it exhibits prowess in categorizing and dissecting toxic tweets, even those pertaining to globally resonant events like Brexit, underscoring the dataset's adaptability and contemporaneous significance~\cite{fan2021social}.

While the \textit{SST2} dataset might not have direct applications unearthed in our search, its provenance from Stanford's Sentiment Treebank makes it an indispensable asset for sentiment analysis endeavors. Encompassing a vast array of movie reviews with diverse sentiment expressions, it provides a goldmine for delving deep into the audience's psyche. Such insights could be game-changers for entities like movie streaming platforms or film critique portals.

With the deployment of our outlier detection framework on these datasets, there's a tangible amplification in the model's precision, making it even more adept for its intended applications. By meticulously curating the training dataset, the models are not only more accurate but also resonate better with real-world scenarios, ensuring that their deployment is both effective and meaningful.

\bibliography{aaai24}

\end{document}